\documentclass[10pt, conference]{IEEEtran}
\IEEEoverridecommandlockouts

\usepackage{adjustbox}
\usepackage{subcaption}
\usepackage{caption}
\usepackage{graphicx}
\usepackage{stfloats}
\usepackage{booktabs}
\usepackage{multirow}

\usepackage[percent]{overpic}
\usepackage{amsmath,amssymb,amsfonts}
\usepackage{float}
\usepackage{multicol}
\usepackage{textcomp}
\usepackage{xcolor}
\usepackage{breakurl}
\usepackage{xurl}
\usepackage{longtable}
\usepackage{array}
\usepackage{verbatim}
\usepackage[ruled, vlined, linesnumbered, boxed]{algorithm2e}
\usepackage{algpseudocode}

\graphicspath{{figures/}}

\title{AI or Real: Detecting Partially Altered Videos Under Resource-Constrained Environments\thanks{This material is based upon work supported by the National Science Foundation (NSF) under Award Number CNS-2401928.}}

\author{
Tamoghna Chakraborty, Md Nurul Absur, Sourya Saha, Saptarshi Debroy\\
City University of New York\\ 
Emails: \textit{\{tchakraborty,mabsur,ssaha2\}@gradcenter.cuny.edu, saptarshi.debroy@hunter.cuny.edu}}

\begin{document}

\maketitle
\thispagestyle{empty}
\pagestyle{empty}
\maketitle

% ---------- Body ----------
\begin{abstract}
The proliferation of generative video models has shifted the practical detection threat from fully fabricated clips to partially manipulated footages. 
%in which a small fraction of frames are AI-generated, preserving scene and context while altering meaning. 
Although modern detectors achieve strong accuracy %on this task 
using foundation backbones of 400M+ parameters, their resource footprint precludes edge deployment.
%on edge hardware where streaming content moderation must occur.
In this paper, we present a lightweight full-frame detector for partially manipulated AI-generated video, designed for deployment on edge hardware without face-detection preprocessing. The system distills a DINOv2-Base teacher into a frozen MobileNetV3-Small student through a pipeline that combines temperature-annealed soft-label transfer, attention-diversity regularization, frame-level supervision, and a residual feature adapter that conditions ImageNet features for artifact detection. We additionally target two failure modes specific to the partial-manipulation regime: false positives on legitimate scene cuts, addressed through within-video temporal hard negatives; and threshold-level miscalibration on the dominant pure-real class, addressed through calibration-aware sampling.
Evaluation on 
%In this evaluation, we test on the GenVidBench Pair2 benchmark with semantically consistent real-fake pairs, constructing 
a 55,393-sample spliced test set across fake-frame ratios from 6.2\% to 31.2\% demonstrates the student model 
%The standalone lightweight baseline- 
closing 58\% of the gap to the DINOv2-Base teacher (AUC 0.766) while running at 3.65 ms per 16-frame clip on RTX A4000 with a 150.4 MB checkpoint compatible with edge-device memory and latency budgets.
\end{abstract}
\section{Introduction}
\label{sec:introduction}

Diffusion-based text-to-video systems now produce photorealistic clips at consumer cost, and the operationally relevant attack pattern is no longer a fully fabricated video but a real clip with a few generated frames inserted, which preserves scene, subject, and context while targeting the threshold of plausibility~\cite{sec2025}. Detecting these manipulations at single-digit fake-frame ratios on resource-constrained edge hardware is a deployment challenge that requires investigation.

State-of-the-art detectors such as UNITE~\cite{kundu2025unite} and STALL~\cite{stall} rely on foundation backbones of $\sim$400M parameters that exceed the memory budget of every edge accelerator shipping at consumer cost, and existing lightweight approaches either require face detection~\cite{lightfakedetect} or predate diffusion-era generators~\cite{rossler2019faceforensics}. No prior work has demonstrated edge-deployable, face-free, partial-fake video detection at streaming-moderation resolution and latency.

In this paper, we present a lightweight detector for partially manipulated AI-generated video that closes this gap. Our approach distills detection capability from a DINOv2-Base teacher~\cite{oquab2023dinov2} into a frozen MobileNetV3-Small~\cite{howard2019searching} student, paying the teacher's compute cost once, offline, and running only the lightweight student at inference. A residual feature adapter reconditions the ImageNet-pretrained student features toward diffusion-artifact statistics; attention-diversity regularization prevents the twelve spatial attention heads from collapsing onto a single region; and a calibration-aware sampling scheme corrects the false-positive shortcut that arises when the model learns to flag scene cuts rather than generative artifacts. A max-frame inference rule preserves sensitivity at low fake-frame ratios by flagging a clip if any single frame's score exceeds threshold. Our contributions are: (i) a knowledge distillation (KD)~\cite{hinton2015kd} pipeline that recovers 58\% of the teacher-student gap while retaining the frozen MobileNetV3 backbone at inference; (ii) two mechanisms for the partial-manipulation regime (within-video hard negatives against scene-cut false positives, calibration-aware sampling for threshold stability); and (iii) the empirical isolation of \textit{feature spatial resolution} as the binding constraint on lightweight detection.

We evaluate the proposed detector on GenVidBench Pair2~\cite{ni2025genvidbench}, a benchmark of semantically consistent real-fake video pairs with fake-frame ratios spanning 6.2\% to 31.2\%. The detector reaches $0.672$ AUC and $50.6\%$ detection at the hardest $6.2\%$-fake tier, up from $0.544$ and $15.8\%$ for a standalone MobileNetV3 baseline. Across four MobileNetV3-S configurations, AUC saturates near $0.67$, while a structurally identical head on $16 \times 16$-resolution DINOv2 features reaches $0.766$, providing empirical evidence that feature spatial resolution is the binding constraint on lightweight detection in this regime.

The remainder of this paper is organized as follows. Section~\ref{sec:related} reviews the related work. Section~\ref{sec:background_and_problem} formalizes the three properties of the partial-manipulation regime. Section~\ref{sec:solution_design} describes the detector architecture, training pipeline, and ablation conditions. Section~\ref{sec:experiments} presents the experimental evaluation. Section~\ref{sec:conclusion} concludes the paper.
\section{Related Work}
\label{sec:related}

Prior work on AI-generated video detection has evolved through three phases: face-centric detection, full-frame detection with foundation backbones, and training-free statistical scoring. Table~\ref{tab:related_comparison} summarizes how representative methods sit along the dimensions relevant to our problem: full-frame operation, partial-manipulation capability, and edge deployability. Below we review each category briefly, focusing on the specific limitation this work addresses.

\begin{table}[t]
\centering
\caption{Comparison of related methods across the dimensions most relevant to our problem. ``Edge'' denotes deployability on edge-class hardware ($\sim$4~GB shared memory).}
\label{tab:related_comparison}
\resizebox{\columnwidth}{!}{%
\begin{tabular}{@{}lccccc@{}}
\toprule
\textbf{Method} & \textbf{Face-} & \textbf{Partial} & \textbf{Edge} & \textbf{Backbone} & \textbf{Params} \\
                & \textbf{free}  & \textbf{fake}    & \textbf{deploy} &                  & \\
\midrule
LightFakeDetect~\cite{lightfakedetect} & \texttimes & \texttimes & $\sim$         & MobileNet-v1     & 22.8M \\
UNITE~\cite{kundu2025unite}            & \checkmark & \texttimes & \texttimes     & SigLIP-So400M    & 400M+ \\
STALL~\cite{stall}                     & \checkmark & \texttimes & \texttimes     & DINOv2           & 86M+  \\
\textbf{Ours (KD Final)}               & \checkmark & \checkmark & \checkmark     & MobileNetV3-S    & \textbf{39.4M} \\
\bottomrule
\end{tabular}%
}
\vspace{-0.2in}
\end{table}

\noindent\textbf{Face-centric detection.}
%\textcolor{red}{
MesoNet~\cite{mesonet} and LightFakeDetect~\cite{lightfakedetect} operate on cropped facial regions extracted by an MTCNN~\cite{mtcnn} front end. LightFakeDetect reaches 99.0\% F1 on Celeb-DF v2 with 22.8M parameters, but the mandatory face-detection stage makes it inapplicable to scenes without human subjects, to background manipulations, or to fully AI-generated non-human content. We remove this constraint by operating on full-frame spatial features throughout, without any face-detection preprocessing.
%}

\noindent\textbf{Full-frame detection with foundation backbones.}
%\textcolor{red}{
VideoMAE~\cite{tong2022videomae} and DeMamba~\cite{chen2024demamba} apply large pretrained backbones to whole-frame synthetic-video detection. UNITE~\cite{kundu2025unite} represents the SOTA in this direction, using SigLIP-So400M ($\sim$400M parameters) with an Attention Diversity (AD) loss that regularizes multi-head attention to attend to spatially diverse regions and separates real/fake class centers in the normalized embedding space. All three methods assume a fully synthetic clip as input, and their memory budgets are incompatible with edge deployment. Our work adapts UNITE's AD-loss formulation to a $\sim$100$\times$ smaller MobileNetV3-Small backbone via knowledge distillation from a DINOv2-Base teacher (Section~\ref{sec:solution_design}), and re-targets the resulting system at the partial-manipulation regime UNITE was not designed for.
%}

\noindent\textbf{Training-free scoring.}
%\textcolor{red}{
STALL~\cite{stall} scores frames against a real-video Gaussian model over DINOv2 embeddings, requires no synthetic training data, and generalizes across unseen generators by design. It is, however, calibrated for \textit{fully} AI-generated content: the frame-level distributional signal STALL exploits is diluted in partially manipulated clips, and the min/max aggregation over per-frame scores does not extend naturally to the case where 1--3 of 16 frames are generated. Rather than avoid training, we address partial manipulation through paired spliced training and a max-frame inference rule that triggers on any single sufficiently anomalous frame.
%}

{\em To the best of our knowledge, this is the first system to simultaneously target full-frame generative-artifact detection, partial manipulation in semantically consistent real-fake pairs, and strict edge deployment within a single framework.}
\section{Background and Problem Motivation}
\label{sec:background_and_problem}

This section formalizes three properties of the partial-manipulation regime that drive our design choices (Section~\ref{sec:solution_design}): the temporal dilution of the detection signal, the elimination of semantic shortcuts in modern paired benchmarks, and the spatial-resolution bottleneck imposed by edge-deployable backbones.

\subsection{Temporal Signal Dilution}
\label{sec:bg_partial}

Detectors trained to distinguish real from fully AI-generated video can exploit global scene-level statistics, aggregate color distributions, object semantics, and clip-wide distributional shifts that vanish when only a small subset of frames in an otherwise real clip has been replaced. At $6$--$20\%$ fake content the temporal signal is dominated by real footage and these global shortcuts fail entirely. The deployment-critical case sits at the lower end of this range: $1$ in $16$ frames generated ($6.2\%$ of the clip), where the detector must integrate weak per-frame evidence across the sequence rather than rely on aggregate statistics. %\textcolor{red}{
This property drives our per-frame supervision and max-frame inference rule.
%}

\subsection{Semantic Consistency}
\label{sec:bg_semantic}

Prior partial-manipulation datasets often used semantically inconsistent splicing, interleaving real footage with generated content from an unrelated domain~\cite{rossler2019faceforensics}. In this evaluation, a simple scene-change detector scores highly by exploiting discontinuities at splice boundaries without identifying any generative artifacts. In-the-wild attacks, in contrast, are specifically crafted to be contextually consistent with the surrounding real footage.

The GenVidBench Pair2 benchmark~\cite{ni2025genvidbench} enforces semantic consistency by providing matched real-fake pairs: for each real clip in HD-VG-130M~\cite{wang2023hdvg}, four AI-generated counterparts are produced by CogVideo, SVD, Mora, and MuseV from the same text prompt and source frame. Real and fake clips therefore depict the same scene, and detection must rely on low-level texture, motion, and temporal-consistency artifacts introduced by the generative pipeline. We adopt Pair2 as our primary evaluation benchmark.

\subsection{The Feature Resolution Bottleneck}
\label{sec:bg_resolution}

Lightweight backbones such as MobileNetV3-Small~\cite{howard2019searching} operate at $7 \times 7$ output spatial resolution, compressing a $224 \times 224$ input frame to a $49$-element grid in which each cell summarizes a $32 \times 32$-pixel patch. At this granularity, individual frames are represented by coarse spatial summaries that pool away the high-frequency texture where diffusion artifacts most reliably appear: periodic noise patterns, subtle boundary blurring, and inconsistencies in film-grain statistics.

%\textcolor{red}{
The problem is particularly acute for image-to-video generators. SVD conditions its generation directly on a real source frame, producing content whose coarse spatial features are nearly identical to those of the real input. At $7 \times 7$ resolution the discriminating signal is not present in the representation, and no head architecture or training procedure on top of frozen features can recover it.
%}

Vision Transformers operating at finer patch granularity address this limitation. DINOv2-Base (ViT-B/14)~\cite{oquab2023dinov2} produces $16 \times 16$ patch tokens of dimension 768 for a $224 \times 224$ input, with each token summarizing a $14 \times 14$-pixel patch and retaining the local structure where artifacts manifest. DINOv2-Base, however, requires approximately 4 GB of GPU memory during inference, saturating the shared-memory budget of edge-class accelerators shipping at consumer cost and rendering it undeployable as a stand-alone detector. This tension, high-resolution features are needed for detection but foundation-scale backbones exceed the edge memory budget, motivates the knowledge-distillation approach of Section~\ref{sec:solution_design}. %\textcolor{red}{
Section~\ref{sec:experiments} confirms the ceiling empirically.%}
% Note: no red highlighting in this file. The changes to Section IV are almost
% entirely deletions (redundant table-echoing prose, inline results previews),
% which don't show up in a red-mark scheme. Content that remained is essentially
% unchanged from the original.

\section{Detector Design}
\label{sec:solution_design}

Our detection system, the \textit{KD Student}, retains the deployment footprint of a MobileNetV3-Small backbone while inheriting representational richness from an offline-distilled DINOv2-Base teacher. Below, we describe the dataset, architecture, training objectives, and distillation pipeline.

\subsection{Dataset and Spliced Test Construction}
\label{sec:datasets}

We adopt the \textbf{GenVidBench Pair2} benchmark~\cite{ni2025genvidbench}, in which each real clip from HD-VG-130M~\cite{wang2023hdvg} is paired with four AI-generated counterparts produced by CogVideo~\cite{hong2022cogvideo}, SVD~\cite{blattmann2023stable}, Mora~\cite{yuan2024mora}, and MuseV~\cite{zheng2024musev} from the same prompt and source frame. We identify 13,369 source-video IDs present across all five sources and partition at the source-ID level using an 80/20 split (seed 42), yielding 10,695 train and 2,683 test IDs. We construct partial-fake test samples by replacing $k$ contiguous frames out of $N_f = 16$ in a real clip with the corresponding frames from one of its paired fake clips, for $k \in \{1, \ldots, 5\}$, yielding five fake-ratio tiers spanning 6.2\% to 31.2\% with 10,542 spliced samples per tier ($N = 55{,}393$ total including pure-real samples). The 6.2\% tier is the deployment-critical case.

\paragraph{Decoupled feature extraction.}
Because the backbone is frozen and processes frames independently, we pre-extract per-frame spatial feature maps once and operate on cached features during training, with feature-level splicing constructed on the fly. Frames are selected by an adaptive transition-score procedure ($N_c = 64$ candidate frames, greedy selection of $N_f = 16$ by adjacent cosine-distance score with minimum index spacing), so that paired clips at each spatial resolution remain temporally aligned. Identical procedures are used for MobileNetV3-S (student) and DINOv2-B (teacher) features.

\subsection{Model Architecture}
\label{sec:architecture}
\label{sec:max_frame}

\begin{figure}[t]
    \centering
    \includegraphics[width=\columnwidth]{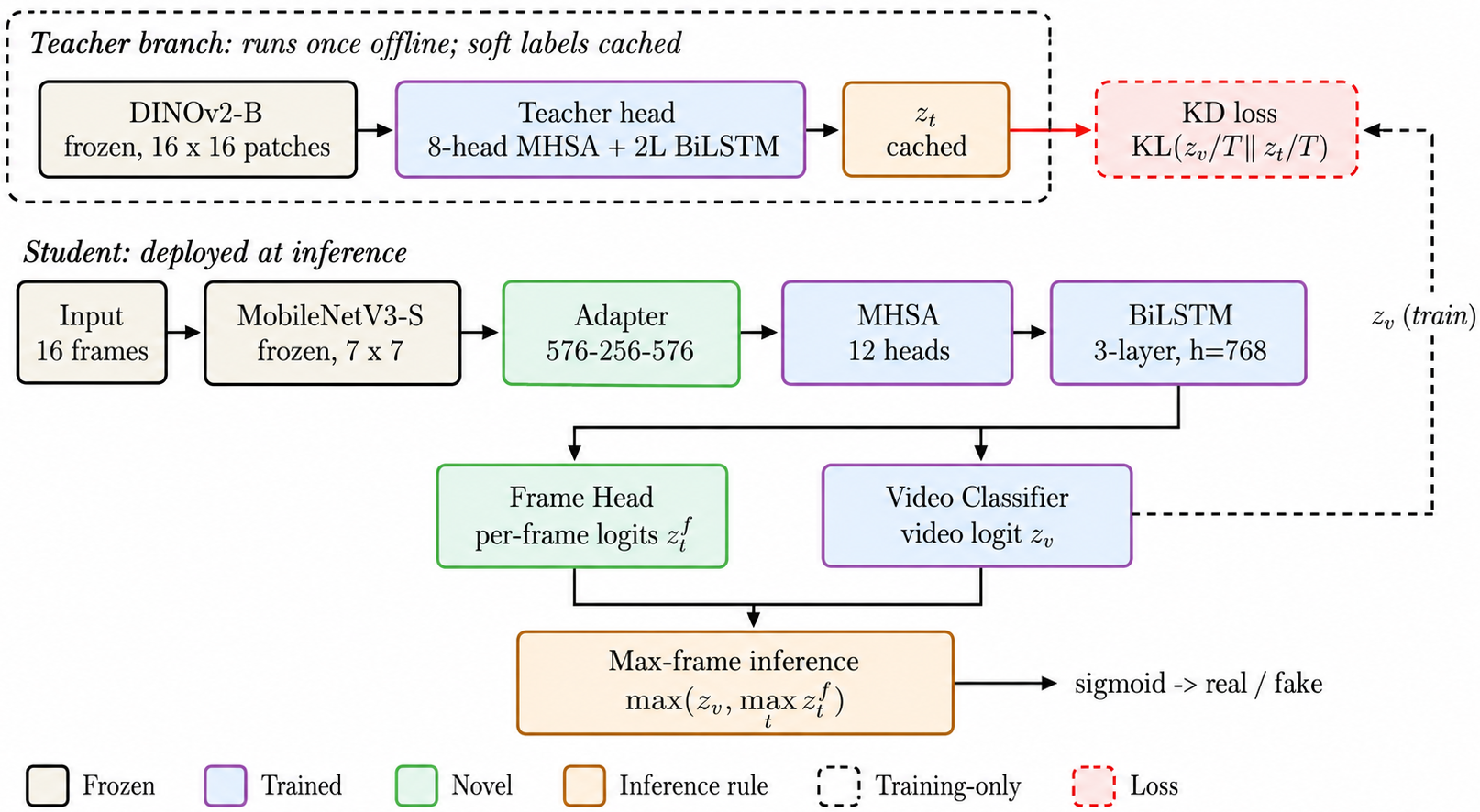}
    \caption{KD Student architecture and knowledge-distillation pathway. The teacher branch (top, dashed) runs once offline; cached teacher logits $z_t$ are combined with student logits $z_v$ in the KD loss during training. At inference, only the student pipeline executes: a frozen MobileNetV3-Small backbone, a residual feature adapter, 12-head spatial attention, and a 3-layer BiLSTM, with the BiLSTM output split between a per-frame head and a video classifier whose logits combine via a max-frame rule. Stars ($\star$) mark novel contributions.}
    \label{fig:architecture}
    \vspace{-0.2in}
\end{figure}

The architecture (Figure~\ref{fig:architecture}) consists of five trained components on top of a frozen MobileNetV3-Small~\cite{howard2019searching}: a \textit{feature adapter}, a \textit{multi-head spatial attention} (MHSA) module, a bidirectional LSTM, a video classifier, and an auxiliary frame head.

\paragraph{Feature adapter.}
\label{sec:adapter}
ImageNet-pretrained features are object-semantic, while generative artifacts manifest in lower-level texture and frequency statistics. We introduce a residual MLP adapter that conditions each frame's globally pooled feature: given $\mathbf{p} = \mathrm{AvgPool}(\mathbf{F}) \in \mathbb{R}^{576}$, we compute $\tilde{\mathbf{F}} = \mathbf{F} + \mathrm{Broadcast}[\mathrm{MLP}(\mathbf{p})]$, where the $576 \to 256 \to 576$ MLP uses LayerNorm and GELU. The output projection is zero-initialized so the adapter starts as identity, allowing distillation to begin without distorting the student's initial calibration. The adapter adds $\sim$0.3M parameters.

\paragraph{Spatial attention, temporal aggregation, and inference.}
The MHSA module uses $K = 12$ heads, each producing a softmax-normalized attention map via a $1 \times 1$ convolutional bottleneck (reduction $r = 4$); per-head attended descriptors are mean-reduced to a per-frame embedding, and per-head attention maps are retained as auxiliary tensors used by the AD loss (Section~\ref{sec:training_objectives}). The 16-frame embeddings are passed to a 3-layer bidirectional LSTM (hidden 768, dropout 0.3); the final hidden state is projected to a video logit $z_v$ through a $1{,}536 \to 768 \to 256 \to 1$ classifier with LayerNorm and GELU. An auxiliary \textbf{frame head} produces per-frame logits $z_t^{(f)}$ from each frame's pooled embedding. The deployed prediction combines them via a max-frame rule, flagging the video if either the global classifier or any single frame is suspicious. The rule being $\hat{p} = \sigma(\max[z_v, \max_t z_t^{(f)}])$.

The full system has 39.43M trainable parameters and a 150.4~MB checkpoint. Inference latency on RTX A4000 is 3.65~ms per 16-frame clip; estimated edge-device (Jetson-class) latency is 139--240~ms.

\subsection{Training Objectives}
\label{sec:training_objectives}
\label{sec:calibration}
\label{sec:hard_negatives}

The KD Student is trained on a virtual dataset of 24,000 samples per epoch drawn from four sample types: spliced real-fake pairs (55\%), pure real (25\%), pure fake (10\%), and within-video temporal hard negatives (10\%). Because the backbone is frozen, feature-level splicing is mathematically equivalent to video-level splicing; the 1/16 tier is oversampled $5\times$ within the spliced fraction to bias optimization toward the deployment-critical case. The replaced frame indices serve as ground truth for the frame head.

\paragraph{Calibration sampling: a deliberate AUC--FPR trade-off.}
A central design choice is the deliberate trade-off between raw discrimination and threshold-level operating characteristics. Earlier configurations with a low pure-real sampling fraction ($\sim$8\%) produced higher headline AUC and detection rates above 80\% at the 6.2\% tier, coupled with FPR exceeding $0.80$ at threshold $0.5$. Probing this configuration with within-video real-real splices revealed that the model triggered on any temporal discontinuity, not specifically on generative artifacts. Increasing the pure-real sampling fraction to 25\% reduced FPR at threshold $0.5$ from above $0.80$ to $0.17$, at the cost of reducing detection at $6.2\%$ from above 80\% to 50.6\%. A content-moderation pipeline that flags one in five real clips is unusable regardless of the detection rate; we treat the calibrated configuration as the canonical one. All results in this paper are from this configuration.

\paragraph{Within-video hard negatives.}
To prevent the model from learning to flag any temporal discontinuity, including legitimate scene cuts in real video, 10\% of training samples are constructed as \textbf{within-video hard negatives}: a single real video is split at a random index, the halves reordered, and the result labeled \textit{real}. The clip contains a real temporal cut but no generative artifact, supplying an unambiguous training signal that temporal discontinuity alone does not justify a positive decision.

\paragraph{Combined objective.}
We use binary focal loss~\cite{lin2017focal} ($\gamma = 2.0$) for both video and frame heads. The KD loss is the temperature-scaled KL divergence
\begin{equation}
    \mathcal{L}_{\text{KD}} = T^2 \cdot \mathrm{KL}\!\left(\sigma(z_s/T) \,\|\, \sigma(z_t/T)\right),
\end{equation}
with temperature annealed linearly from $T = 6.0$ to $T = 3.0$~\cite{hinton2015kd}. For spliced samples, the teacher logit is a fake-ratio interpolation $z_t^{\text{splice}} = (1 - r_{\text{fake}}) z_t^{\text{real}} + r_{\text{fake}} z_t^{\text{fake}}$. The AD loss~\cite{kundu2025unite} regularizes head-averaged $\ell_2$-normalized embeddings $\hat{\mathbf{e}}_i$ via within-class compactness and between-class separation terms:
\begin{align}
    \mathcal{L}_{\text{within}}  &= \tfrac{1}{|\mathcal{C}|}\!\sum_{c}\, \mathrm{mean}_{i: y_i=c}[\mathrm{ReLU}(\delta_w - \hat{\mathbf{e}}_i \cdot \hat{\boldsymbol{\mu}}_c)], \\
    \mathcal{L}_{\text{between}} &= \mathrm{ReLU}(\delta_b - (1 - \hat{\boldsymbol{\mu}}_{\text{real}} \cdot \hat{\boldsymbol{\mu}}_{\text{fake}})),
\end{align}
with $\delta_w = 0.3$, $\delta_b = 0.5$, and class means computed in-batch. A complementary term penalizes cosine similarity between attention maps across the 12 heads. The combined loss is $\mathcal{L} = 0.40 \mathcal{L}_{\text{focal}}^{(v)} + 0.35 \mathcal{L}_{\text{KD}} + 0.25 \mathcal{L}_{\text{focal}}^{(f)} + 0.15 (\mathcal{L}_{\text{within}} + \mathcal{L}_{\text{between}}) + 0.05 \mathcal{L}_{\text{div}}$.

\subsection{Distillation Pipeline and Ablation Conditions}
\label{sec:kd_pipeline}

The teacher (DINOv2-Base~\cite{oquab2023dinov2}, 8-head attention, 2-layer BiLSTM hidden 256) is trained on Pair2 with BCE and AD loss and produces cached soft labels; the student is trained against these soft labels, with no live teacher forward pass during training and no teacher dependency at inference. We evaluate four MobileNetV3-S configurations and the teacher as an upper-bound reference (Table~\ref{tab:ablation}). The Lightweight Baseline uses 4-head MHSA and a single-layer BiLSTM, trained with BCE and spliced samples only; KD Fine-tuned additionally unfreezes the last three convolutional blocks of the backbone with differential learning rates ($10^{-5}$ backbone, $3 \times 10^{-5}$ head). All students are trained for 30 epochs of 800 steps at batch size 32, using AdamW at $3 \times 10^{-5}$ with cosine annealing and warm restarts ($T_0 = 10$).

\begin{table}[b]
\centering
\footnotesize
\caption{Ablation conditions. All MobileNetV3-S configurations share the frozen ImageNet backbone, adaptive frame selection, and Pair2 split. The DINOv2-B teacher is shown as an upper-bound reference. ``BB FT'' = backbone fine-tuning.}
\label{tab:ablation}
\setlength{\tabcolsep}{4pt}
\renewcommand{\arraystretch}{1.05}
\begin{tabular}{@{}lccccc@{}}
\toprule
\textbf{Model} & \textbf{KD} & \textbf{Adapter} & \textbf{AD} & \textbf{BB FT} & \textbf{Edge} \\
\midrule
Lightweight Baseline           & \texttimes & \texttimes & \texttimes & \texttimes & \checkmark \\
KD w/o Adapter                 & \checkmark & \texttimes & \checkmark & \texttimes & \checkmark \\
\textbf{KD Final (proposed)}   & \checkmark & \checkmark & \checkmark & \texttimes & \checkmark \\
KD Fine-tuned                  & \checkmark & \checkmark & \checkmark & \checkmark & \checkmark \\
DINOv2-B Teacher$^\dagger$     & \texttimes & \texttimes & \checkmark & \texttimes & \texttimes \\
\bottomrule
\end{tabular}
\end{table}
\section{Evaluation and Results}
\label{sec:experiments}

We evaluate the KD Student on the GenVidBench Pair2 spliced test split (Section~\ref{sec:datasets}) against four reference points: a Lightweight Baseline that fixes the standalone MobileNetV3 ceiling, a KD-without-Adapter ablation that isolates the contribution of the feature adapter (described in Section~\ref{sec:architecture}), a KD Fine-tuned variant that unfreezes the last three backbone blocks, and the DINOv2-B teacher as an upper-bound reference. %\textcolor{red}{
The five configurations are summarized in Table~\ref{tab:ablation}.%} 
AUC is the primary metric; detection rate and FPR at threshold $0.5$ are reported together throughout. %\textcolor{red}{
Table~\ref{tab:setup} summarizes the experimental setup.
%}

%{\color{red}
\begin{table}[b]
\centering
\scriptsize
\caption{Experimental setup. Training and evaluation are on a single NVIDIA RTX A4000 (16~GB).}
\label{tab:setup}
\setlength{\tabcolsep}{4pt}
\renewcommand{\arraystretch}{1.12}
\begin{tabular}{@{}ll@{}}
\toprule
\textbf{Aspect} & \textbf{Setting} \\
\midrule
Benchmark               & GenVidBench Pair2 \\
Total test samples      & 55{,}393 \\
Test IDs                & 2{,}683 (seed 42, 80/20 by source ID) \\
Splices per ID          & 4 generators $\times$ 5 fake-ratio tiers \\
Fake ratios             & 1/16, 2/16, 3/16, 4/16, 5/16 (6.2\%--31.2\%) \\
\midrule
Frames per clip         & 16 \\
Frame selection         & Adaptive transition-score ($N_c = 64$) \\
Optimizer               & AdamW, LR $3 \times 10^{-5}$, cosine warm restarts ($T_0 = 10$) \\
Epochs / batch          & 30 epochs of 800 steps, batch 32 \\
Backbone (KD Fine-tuned)& Last 3 blocks unfrozen at LR $10^{-5}$ \\
\midrule
Inference rule          & Max-frame: $\max(z_v, \max_t z_t^{(f)})$ \\
Metrics                 & AUC, F1, FPR@0.5, Det@fake-ratio \\
Latency                 & Measured at batch $=$ 1; edge estimated from compute ratio \\
\bottomrule
\end{tabular}
%\vspace{-0.2in}
\end{table}
%}

% ============================================================
\subsection{Main Results}
\label{sec:main_results}

\begin{figure}[t]
    \centering
    \includegraphics[width=\linewidth]{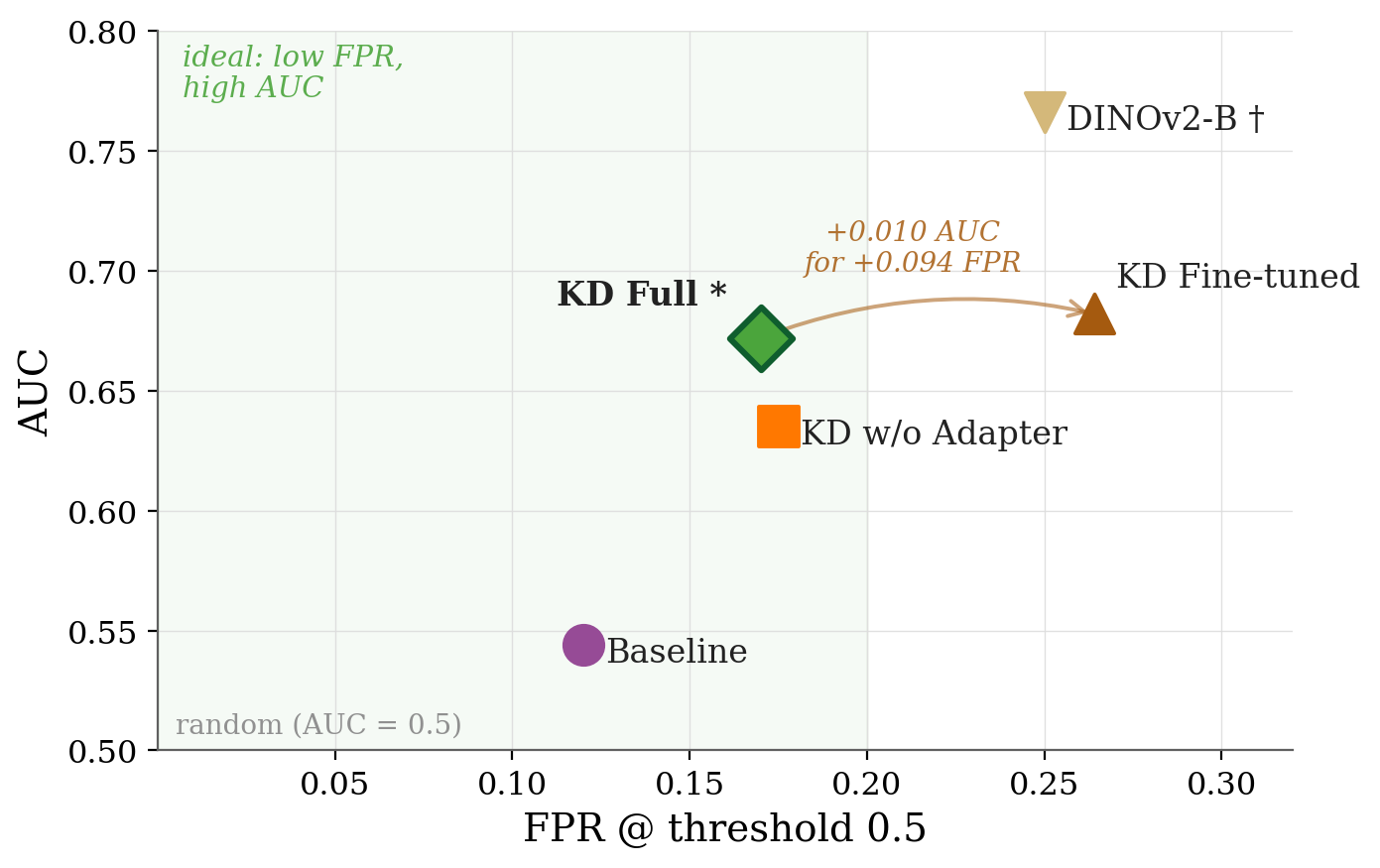}
    \vspace{-0.2in}
    \caption{AUC vs FPR trade-off across all configurations. Better models sit toward the upper-left corner. KD Final at (0.170, 0.672) is the leftmost of the KD-family points; KD Fine-tuned buys +0.010 AUC by paying +0.094 FPR, which we consider the wrong operational trade-off. The DINOv2-B teacher achieves the highest AUC but at an FPR comparable to KD Fine-tuned and outside the edge memory budget.}
    \label{fig:auc_vs_fpr}
    %\vspace{-0.1in}
\end{figure}

Table~\ref{tab:main_results} reports overall metrics and deployment statistics for all five configurations. Table~\ref{tab:contribution} decomposes the AUC and FPR movements as incremental component contributions, isolating in particular the feature adapter (Section~\ref{sec:adapter}); Figure~\ref{fig:auc_vs_fpr} visualizes the same operating points on the AUC-vs-FPR plane.

\begin{table*}[!b]
\vspace{-0.15in}
\centering
\scriptsize
\setlength{\tabcolsep}{4.5pt}
\renewcommand{\arraystretch}{1.15}
\caption{Main results on the GenVidBench Pair2 spliced test split ($N = 55{,}393$). All MobileNetV3-S configurations share the frozen ImageNet backbone and use the max-frame inference rule (Section~\ref{sec:max_frame}). \textbf{KD Final} is the proposed system (frozen backbone + adapter + 12-head MHSA + 3-layer BiLSTM + KD + AD loss + frame supervision + hard negatives + calibration sampling); \textbf{KD w/o Adapter} is the same pipeline with the feature adapter removed; \textbf{KD Fine-tuned} is KD Final with the last three convolutional blocks of the backbone unfrozen and trained with differential learning rates. $^\dagger$ DINOv2-B Teacher is not edge deployable and is shown as an upper-bound reference. AUC is the primary metric; detection rate and FPR are at threshold $0.5$. Latency is measured on RTX A4000 (batch~$=$~1); edge latency is estimated from the compute ratio.}
\label{tab:main_results}
\vspace{-0.05in}
\resizebox{\textwidth}{!}{%
\begin{tabular}{lccccccccc}
\toprule
\multirow{2}{*}{\textbf{Model}} &
\multirow{2}{*}{\textbf{Backbone}} &
\multirow{2}{*}{\textbf{Params}} &
\multirow{2}{*}{\textbf{Size}} &
\multirow{2}{*}{\textbf{AUC}} &
\multirow{2}{*}{\textbf{F1}} &
\textbf{FPR} &
\textbf{Latency} &
\textbf{Edge} &
\multirow{2}{*}{\textbf{Deploy}} \\
& & & & & & \textbf{@0.5} & \textbf{(A4000)} & \textbf{est.} & \\
\midrule
Lightweight Baseline
  & MobileNetV3-S & 1.09M  & 4.16~MB  & 0.544 & 0.311 & 0.120 & 1.00~ms & 40--60~ms & \checkmark \\
KD w/o Adapter
  & MobileNetV3-S & 39.13M & 149.3~MB & 0.635 & 0.622 & 0.175 & 3.50~ms & 130--220~ms & \checkmark \\
\textbf{KD Final (proposed)}
  & MobileNetV3-S & \textbf{39.43M} & \textbf{150.4~MB} & \textbf{0.672} & \textbf{0.690} & \textbf{0.170} & \textbf{3.65~ms} & \textbf{139--240~ms} & \checkmark \\
KD Fine-tuned
  & MobileNetV3-S & 40.07M & 154.1~MB & 0.682 & 0.704 & 0.264 & 3.80~ms & 145--250~ms & \checkmark \\
\midrule
DINOv2-B Teacher$^\dagger$
  & DINOv2-B (ViT-B/14) & 87M & 330~MB & \textit{0.766} & \textit{0.733} & 0.250 & $\sim$80~ms & not deployable & \texttimes \\
\bottomrule
\end{tabular}%
}
%\vspace{-0.2in}
\end{table*}

%{\color{red}
\begin{table}[!b]
\centering
\scriptsize
\caption{Ablation contribution. Each row shows the incremental change from adding one component to the previous configuration. Backbone unfreezing yields the smallest AUC gain and the largest FPR degradation.}
\label{tab:contribution}
\setlength{\tabcolsep}{5pt}
\renewcommand{\arraystretch}{1.15}
\begin{tabular}{lccc}
\toprule
\textbf{Component added} & $\boldsymbol{\Delta}$ \textbf{AUC} & $\boldsymbol{\Delta}$ \textbf{Det@6.2\%} & $\boldsymbol{\Delta}$ \textbf{FPR@0.5} \\
\midrule
KD pipeline (vs.\ Baseline)             & +0.091 & +27.0 pp & +0.055 \\
Feature Adapter (vs.\ KD w/o Adapter)   & +0.037 & +7.8 pp  & $-0.005$ \\
Backbone unfreezing (vs.\ KD Final)     & +0.010 & +1.2 pp  & \textbf{+0.094} \\
\bottomrule
\end{tabular}
%\vspace{-0.1in}
\end{table}
%}

%{\color{red}
%}

\textbf{Knowledge distillation closes 58\% of the teacher-baseline gap.} The Lightweight Baseline reaches 0.544 AUC and the DINOv2-B teacher 0.766. KD Final reaches 0.672, a 0.128 improvement over the baseline that recovers 58\% of the teacher-baseline gap while retaining the frozen MobileNetV3 backbone at inference.

\textbf{The reported KD Final operating point reflects a deliberate AUC vs.\ FPR trade-off.} An earlier configuration trained with a low pure-real sampling fraction achieved higher headline AUC and 6.2\% detection rates above $80\%$, but produced FPR exceeding $0.80$ at threshold $0.5$, which probing with within-video real-real splices (Section~\ref{sec:realreal}) revealed to be a temporal-discontinuity shortcut rather than artifact-based detection. Re-balancing toward the calibrated configuration (Section~\ref{sec:calibration}) reduced FPR to $0.170$ at the cost of detection at $6.2\%$ dropping to $50.6\%$. A content-moderation pipeline that flags one in five real clips is unusable regardless of ranking quality; we treat the calibrated configuration as canonical, and all numbers reported herein are from it. Figure~\ref{fig:auc_vs_fpr} places KD Final in the upper-left region of the trade-off plane, closer to the ideal (low FPR, high AUC) than any other deployable configuration.

% ============================================================
\subsection{Detection Rate by Fake-Frame Ratio}
\label{sec:detection_rate}

\begin{figure}[t]
    \centering
    \includegraphics[width=\linewidth]{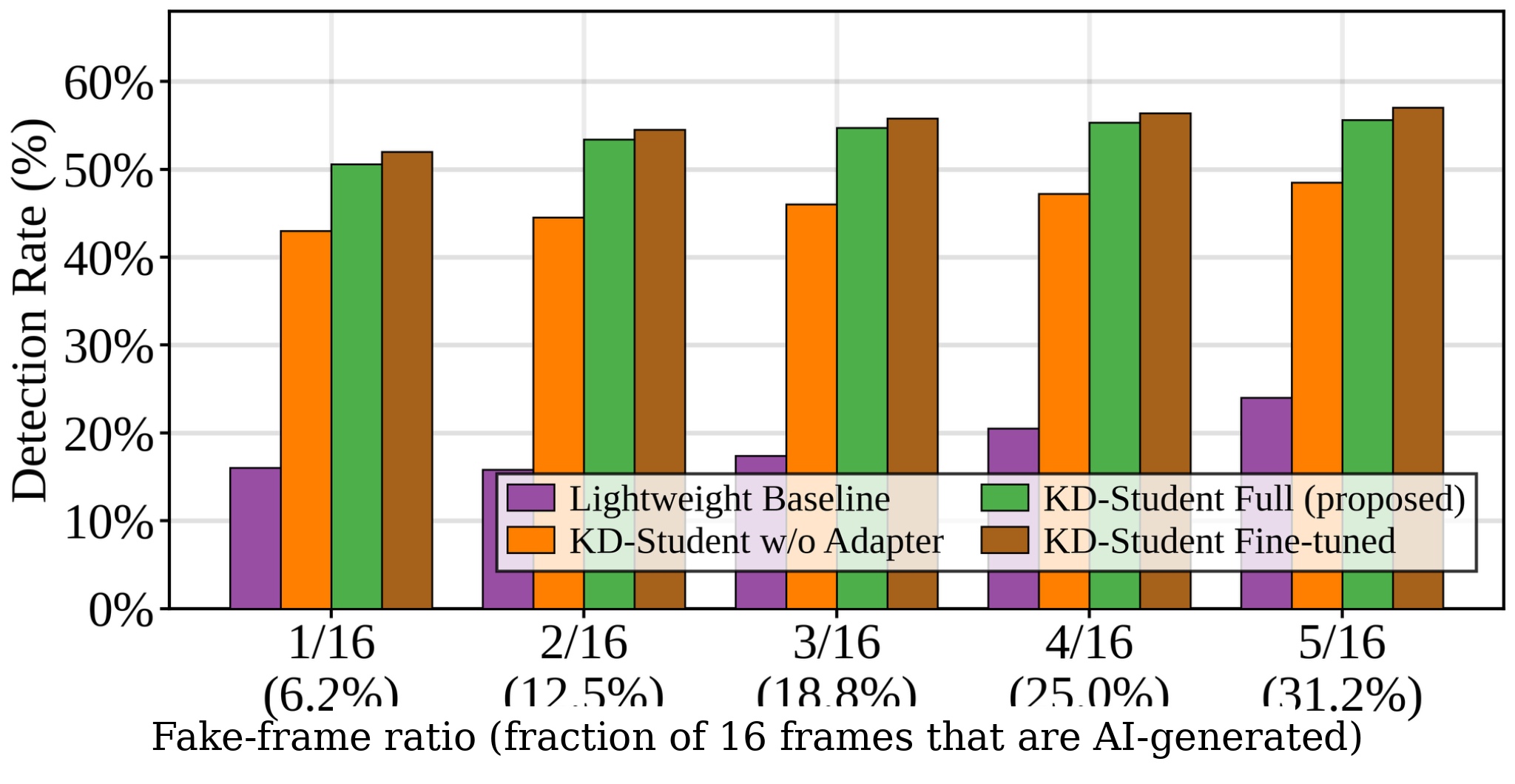}
    \vspace{-0.2in}
    \caption{Detection rate as a function of fake-frame ratio. KD Final reaches 50.6\% detection at the deployment-critical 6.2\% tier with FPR $0.170$, compared to 15.8\% for the Lightweight Baseline at FPR $0.120$. Curves are flatter for the MobileNetV3-S than for the DINOv2-B teacher, reflecting the resolution ceiling on per-frame artifact detection.}
    \label{fig:detection}
    \vspace{-0.2in}
\end{figure}

Table~\ref{tab:detection_rate} and Figure~\ref{fig:detection} report detection rate (TPR at threshold $0.5$) across the five fake-ratio tiers.

\begin{table}[b]
\centering
\scriptsize
\caption{Detection rate (TPR at threshold $0.5$) by fake-frame ratio. FPR at the same threshold is shown alongside. The DINOv2-B teacher is included for completeness but is not directly comparable at this threshold (see Section~\ref{sec:detection_rate}). $n = 10{,}542$ spliced samples per ratio.}
\label{tab:detection_rate}
\setlength{\tabcolsep}{4.5pt}
\renewcommand{\arraystretch}{1.12}
\begin{tabular}{lccccccc}
\toprule
\textbf{Model} & \textbf{1/16} & \textbf{2/16} & \textbf{3/16} & \textbf{4/16} & \textbf{5/16} & \textbf{FPR} \\
& \textbf{(6.2\%)} & \textbf{(12.5\%)} & \textbf{(18.8\%)} & \textbf{(25.0\%)} & \textbf{(31.2\%)} & \textbf{@0.5} \\
\midrule
Lightweight Baseline       & 15.8\% & 15.5\% & 17.4\% & 20.3\% & 23.6\% & 12.0\% \\
KD w/o Adapter             & 42.8\% & 44.5\% & 45.7\% & 47.1\% & 48.2\% & 17.5\% \\
\textbf{KD Final}          & \textbf{50.6\%} & \textbf{53.0\%} & \textbf{54.4\%} & \textbf{55.1\%} & \textbf{55.6\%} & \textbf{17.0\%} \\
KD Fine-tuned              & 51.8\% & 54.3\% & 55.6\% & 56.4\% & 57.2\% & 26.4\% \\
\midrule
DINOv2-B Teacher$^\dagger$ & 39.3\% & 53.5\% & 61.3\% & 66.7\% & 70.9\% & 25.0\% \\
\bottomrule
\end{tabular}
%\vspace{-0.2in}
\end{table}

\textbf{KD Final detects half of single-frame splices at 17\% FPR.} On the 6.2\% tier, KD Final detects 50.6\% of spliced samples at threshold $0.5$, compared to 15.8\% for the Lightweight Baseline. Both operate on identical MobileNetV3 features at inference; the improvement is attributable to the training procedure alone.

\textbf{Detection is roughly flat across the MobileNetV3-S family.} For KD Final, detection rises from 50.6\% (1/16) to 55.6\% (5/16). Under the max-frame inference rule, a positive decision can be triggered by a single sufficiently anomalous frame, so additional fake frames primarily add redundant evidence. The flat curves are consistent with, and partially validate, the max-frame design.

\textbf{Teacher detection at threshold $0.5$ is below KD Final's despite higher AUC, and this is calibration, not capability.} The teacher reaches 0.766 AUC but only 39.3\% detection at the 6.2\% tier at threshold $0.5$. The teacher is trained with standard BCE and produces well-calibrated probabilities, while the student is trained with focal loss, low-ratio oversampling, and calibration sampling, all of which deliberately make it more aggressive at threshold $0.5$. AUC, which is threshold-free, correctly identifies the teacher as the higher-discrimination model.

% ============================================================
\subsection{Per-Generator Discrimination}
\label{sec:per_generator}

\begin{figure}[t]
    \centering
    \includegraphics[width=0.85\linewidth]{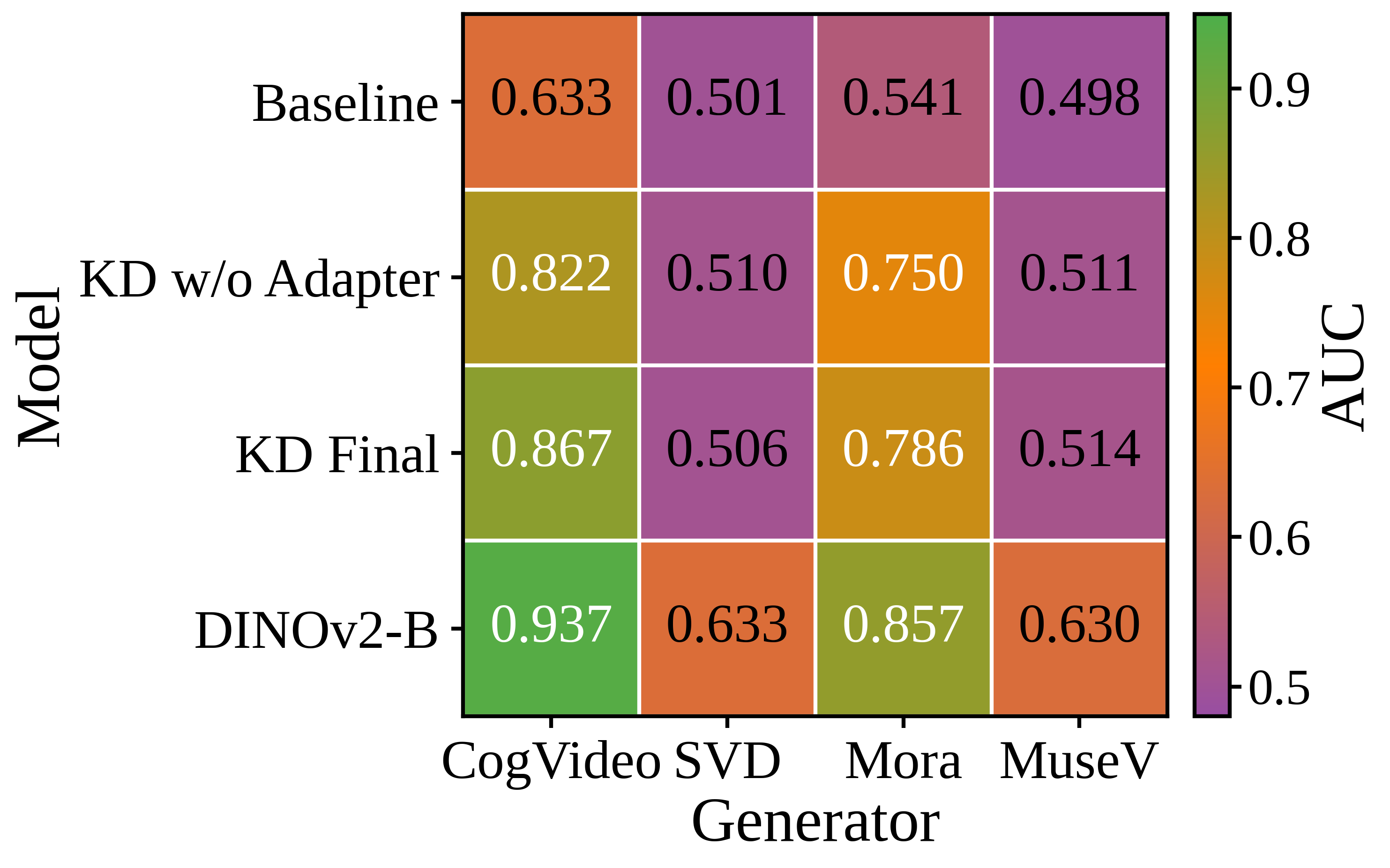}
    \caption{Per-generator AUC on the Pair2 spliced test split. CogVideo and Mora are detectable across configurations as model capacity grows (KD Final: 0.867 and 0.786). SVD and MuseV remain near-random ($0.50 \pm 0.02$) across all MobileNetV3-S models and only reach 0.633 and 0.630 for the DINOv2-B teacher.}
    \label{fig:per_generator}
    \vspace{-0.2in}
\end{figure}

Figure~\ref{fig:per_generator} shows per-generator AUC across all configurations, revealing a consistent two-tier structure. CogVideo (text-conditioned) and Mora (keyframe-based) are detectable across the model family, with KD Final closing most of the gap to the teacher (0.867 and 0.786 vs.\ the teacher's 0.937 and 0.857). SVD and MuseV remain near random for every MobileNetV3-S configuration; the teacher itself only reaches 0.633 and 0.630. SVD conditions its generation directly on real video frames, so the output inherits the appearance statistics of the real source and is featurally close to real even at $16 \times 16$ patch resolution. We interpret SVD and MuseV as a fundamental task-difficulty ceiling for feature-based detection on this benchmark: the artifact signatures these image-to-video generators leave are too weak to be detected from frozen visual features alone, regardless of head architecture or training procedure.

% ============================================================
\subsection{Backbone Fine-Tuning}
\label{sec:finetune}

KD Fine-tuned unfreezes the last three convolutional blocks of MobileNetV3-S with differential learning rates and is trained under the same sampling and loss configuration as KD Final. It reaches 0.682 AUC, a 0.010 improvement over KD Final at the cost of FPR rising from $0.170$ to $0.264$ (a 55\% relative increase) for a 1.5\% relative AUC gain. Per-generator pattern, ratio dependence, and ceiling near 0.68 are otherwise indistinguishable from the frozen configuration. Under the calibration argument of Section~\ref{sec:calibration}, this is the wrong trade-off (see also Figure~\ref{fig:auc_vs_fpr}); we retain KD Final as the recommended deployable system.

\paragraph{Bypass-validation caveat.}
Validation during fine-tuning was conducted on features pre-extracted from the \textit{original} frozen backbone, i.e., feature extraction bypassed the very blocks being fine-tuned. Bypass-validation AUC consequently reached 0.776, suggesting strong improvement, while full raw-video evaluation places the same checkpoint at 0.682. The bypass figure is uninformative about deployment behavior. We highlight this as a cautionary note for two-stage training pipelines: validation must use the same forward pass as deployment, not a feature-cached approximation.

% ============================================================
\subsection{Computational Footprint and Edge Deployability}
\label{sec:efficiency}

\begin{figure}[t]
    \centering
    \includegraphics[width=\linewidth]{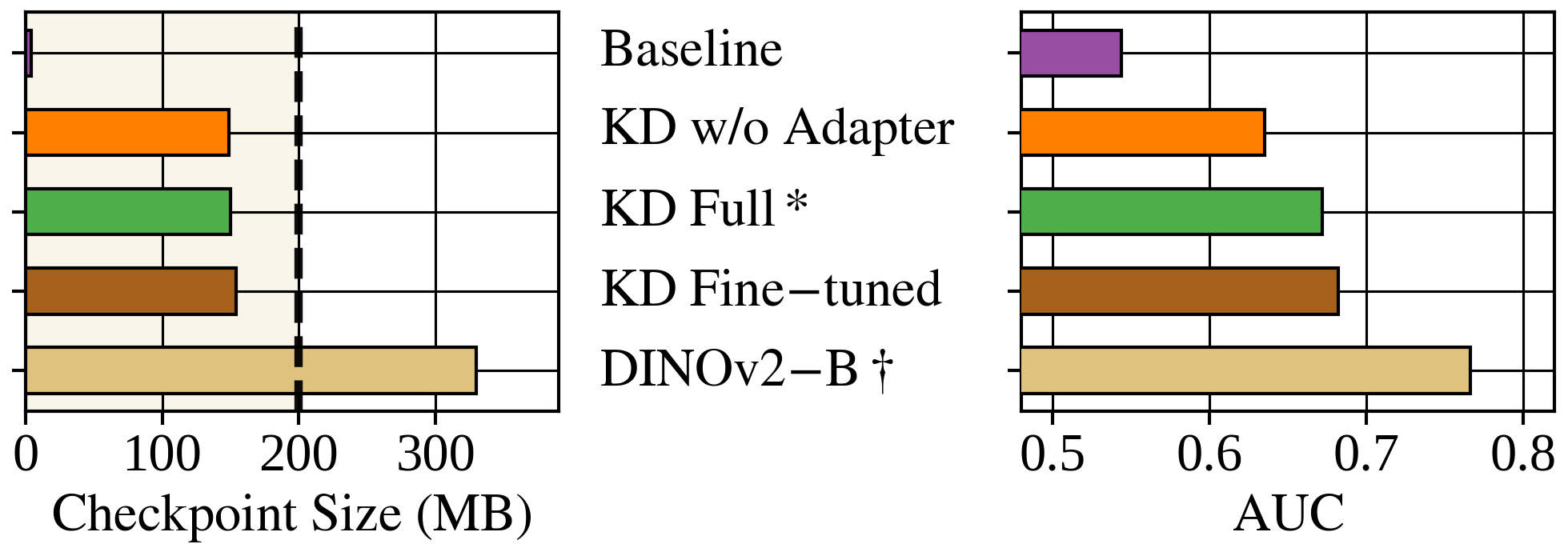}
    \caption{Efficiency trade-off between checkpoint size and detection performance. The left panel reports checkpoint size, where the dashed vertical line marks the 200~MB edge-deployable budget and the shaded region denotes models within this budget. The right panel reports the AUC, with the dashed vertical line marking the random baseline.}
    \label{fig:efficiency}
    \vspace{-0.2in}
\end{figure}

KD Final's footprint (39.43M trainable parameters, 150.4~MB checkpoint, 3.65~ms RTX A4000 latency per 16-frame clip, 352~MB peak VRAM) places it inside the deployable region of Figure~\ref{fig:efficiency}. Estimated edge-class (Jetson-scale) latency, computed from a Maxwell-vs-Ampere compute ratio with a memory-bandwidth derating for the LSTM, falls in the 139--240~ms range per clip, within a one-clip-per-second real-time inference budget. This estimate is computed from architectural ratios and is not measured on hardware; we report it with that qualification. The DINOv2-B teacher, at 330~MB with $\sim$4~GB peak VRAM, saturates the shared-memory budget of edge-class accelerators shipping at consumer cost and is not deployable on the target hardware; we include it only as an upper bound on what richer features could achieve under the same training procedure.

% ============================================================
\subsection{Robustness to Genuine Temporal Discontinuity}
\label{sec:realreal}

A practical concern for any partial-fake detector is whether the model has learned to detect generative artifacts or has shortcutted to detecting any temporal discontinuity, including legitimate scene cuts in real video. The within-video hard negatives in training (Section~\ref{sec:hard_negatives}) target this failure mode directly. To probe whether they succeed, we constructed four pairs of within-video real-real splices from held-out YouTube clips, in which two halves of a single real video are reordered to create a temporal cut without any generated content; we additionally compared against four real-fake spliced clips from the same source IDs.

Across the four real-real splice configurations, KD Final produced predicted probabilities in the range $0.51$--$0.55$, with one of four crossing threshold $0.5$ (an effective FPR of 25\% on this small probe). The same model produced predicted probabilities of $0.87$ or higher on the real-fake comparison clips, a separation of approximately $0.32$. We interpret this as preliminary evidence that the model is detecting generative artifacts rather than scene cuts per se, while noting that the probe size ($n = 4$ per condition) does not constitute a rigorous statistical claim. A larger-scale evaluation against curated real-real splices is left for future work.

All datasets, feature extraction scripts, and a detailed README are provided as a GitHub repository \cite{git}.
\section{Discussions and Conclusions}
\label{sec:conclusion}

We presented a lightweight detector for partially manipulated AI-generated video, designed for edge-class hardware without face-detection preprocessing. The system distills a DINOv2-Base teacher into a frozen MobileNetV3-Small student using temperature-annealed soft-label transfer, attention-diversity regularization, frame-level supervision, and a residual feature adapter. Two failure modes specific to partial manipulation, false positives on legitimate scene cuts and miscalibration on the dominant pure-real class, are addressed through within-video temporal hard negatives and calibration-aware sampling. On the GenVidBench Pair2 spliced test split, the system achieves $0.672$ AUC and $50.6\%$ detection at the deployment-critical $6.2\%$ fake-frame ratio (FPR $0.170$), closing $58\%$ of the $0.222$ AUC gap to the DINOv2-B teacher while processing a 16-frame clip in $3.65$~ms with a $150.4$~MB checkpoint.

Across four MobileNetV3-S configurations spanning head design, distillation, regularization, and partial backbone fine-tuning, AUC saturates near $0.68$. This remains $0.094$ below the structurally identical head trained on $16 \times 16$-resolution DINOv2 features. We interpret this as evidence that the main constraint on lightweight detection is the spatial resolution of the per-frame representation, rather than the temporal architecture, training procedure, or distillation strategy. A related pattern appears at the generator level: SVD and MuseV remain near random across all configurations, including the teacher, because both condition generation on real source frames and inherit their appearance statistics. Feature-based detection alone is therefore insufficient. Addressing image-conditioned generators may require complementary signals such as perturbation-based temporal consistency, frequency-domain analysis, or external metadata.

\noindent{\textbf{Limitations:}
Three limitations bound these results. First, edge-class latency (139--240~ms) is estimated from compute ratios rather than measured on hardware. Second, the within-video real-real splice probe used to test sensitivity to generative artifacts rather than temporal discontinuity is small ($n = 4$ clip pairs). Third, our backbone fine-tuning experiments initially used a validation procedure that bypassed the trained backbone blocks. Reported fine-tuning results use corrected raw-video evaluation, but earlier training decisions may have been influenced by the bypass. Finally, all results come from a single benchmark, leaving cross-dataset generalization open.

\noindent{\textbf{Future work:}
{Direct validation on a Jetson Nano or comparable device would replace estimated latency and memory costs with measured values. Evaluating higher-resolution lightweight students and alternative foundation teachers would test the resolution-ceiling hypothesis from both directions. A larger curated set of legitimate creator content would strengthen claims about real-world false positives. Cross-dataset evaluation on unseen generators and under practical perturbations would further establish whether the calibrated operating point transfers beyond the training benchmark.}}

Synthetic video detection is increasingly shifting toward partial manipulation in semantically consistent settings and toward resource-constrained edge deployment. The accuracy gap between foundation-model detectors and lightweight alternatives remains substantial, but knowledge distillation, calibration-aware training, and explicit treatment of partial-manipulation failure modes can close part of it. The remaining gap appears structural, arising primarily from feature spatial resolution rather than head architecture or training procedure. This finding identifies feature spatial resolution as a primary target for future edge-deployable detector design.

% ---------- References ----------
\bibliographystyle{ieeetr}
\bibliography{refs}

\end{document}